\documentclass[letterpaper, 10 pt, conference]{ieeeconf}  

\IEEEoverridecommandlockouts                              

\usepackage{graphics} 
\usepackage{subcaption}
\usepackage{hyperref}
\usepackage[ruled,linesnumbered]{algorithm2e}
\usepackage{epsfig} 
\usepackage{mathptmx} 
\usepackage{times} 
\usepackage{amsmath} 
\usepackage{amssymb}  

\title{\LARGE \bf
Learning-based Airflow Inertial Odometry for MAVs using Thermal Anemometers in a GPS and vision denied environment
}

\author{Ze Wang, Jingang Qu, Zhenyu Gao and Pascal Morin
\thanks{Authors are with Institut des Systemes Intelligents et de Robotique - ISIR, Sorbonne University, CNRS, UMR 7222, 75005 Paris, France
        {\tt\small firstName.lastName@isir.upmc.fr}}%
}

\begin{document}

\maketitle
\thispagestyle{empty}
\pagestyle{empty}

\begin{abstract}

This work demonstrates an airflow inertial based odometry system with multi-sensor data fusion, including thermal anemometer, IMU, ESC, and barometer. This goal is challenging because low-cost IMUs and barometers have significant bias, and anemometer measurements are very susceptible to interference from spinning propellers and ground effects. We employ a GRU-based deep neural network to estimate relative air speed from noisy and disturbed anemometer measurements, and an observer with bias model to fuse the sensor data and thus estimate the state of aerial vehicle. A complete flight data, including takeoff and landing on the ground, shows that the approach is able to decouple the downwash induced wind speed caused by propellers and the ground effect, and accurately estimate the flight speed in a wind-free indoor environment. IMU, and barometer bias are effectively estimated, which significantly reduces the position integration drift, which is only 5.7m for 203s manual random flight. The open source is available on \url{https://github.com/SyRoCo-ISIR/Flight-Speed-Estimation-Airflow}.

\end{abstract}

\section{INTRODUCTION} \label{introduction}

Odometry system plays a very important role in mobile robot navigation systems. Depending on the characteristics of different sensors people have designed various odometry systems \cite{Mohamed(2019)}, which are usually categorized into three types, namely external reference odometry systems (GPS, mocap), external context-aware odometry systems (vision, laser), and proprioceptive odometry systems (INS\cite{Herath(2020)}\cite{Liu(2020)}, aerodynamic model \cite{Letalenet(2020)}). The external reference odometry systems and external context-aware odometry systems have high position accuracy, but depend on external conditions and are susceptible to interference, such as signal occlusion, signal reflection, signal jamming, low-light conditions, motion blur, textureless, dynamic environments, etc. Proprioceptive odometry systems are used to obtain position velocity by direct measurement or prediction of system dynamics and thus integration, which can avoid over-dependence on environment and have high sampling rate, but accumulate drift during velocity position derivation due to calibration errors, bias, and noise, and thus usually assists other odometry systems. In low globality scenarios, e.g., teleoperation, the aerial vehicle has a higher dependence on flight speed estimation than position estimation, so in this work, we focus more on speed estimation as well as position drift performance, and a new odometry system avoiding over-reliance on the environment while directly measuring speed.

Wheeled robots typically use electronic encoders to record wheel rotation as an odometer measurement. However, aerial vehicles do not have contact with the ground, instead they have relative air velocity, so we expect to infer the flight speed by measuring the air velocity. Note that the air velocity is the velocity of the vehicle relative to the air at infinity in a constant flow field, which contains the vehicle's velocity and the flow field's velocity.  Fortunately, such interferences usually occur in windless indoor environments. Therefore this work concerns MAVs' odometry system based on airflow sensing in the wind-free environment. 

\begin{figure}[t]
	\centering
	\includegraphics[scale=0.34]{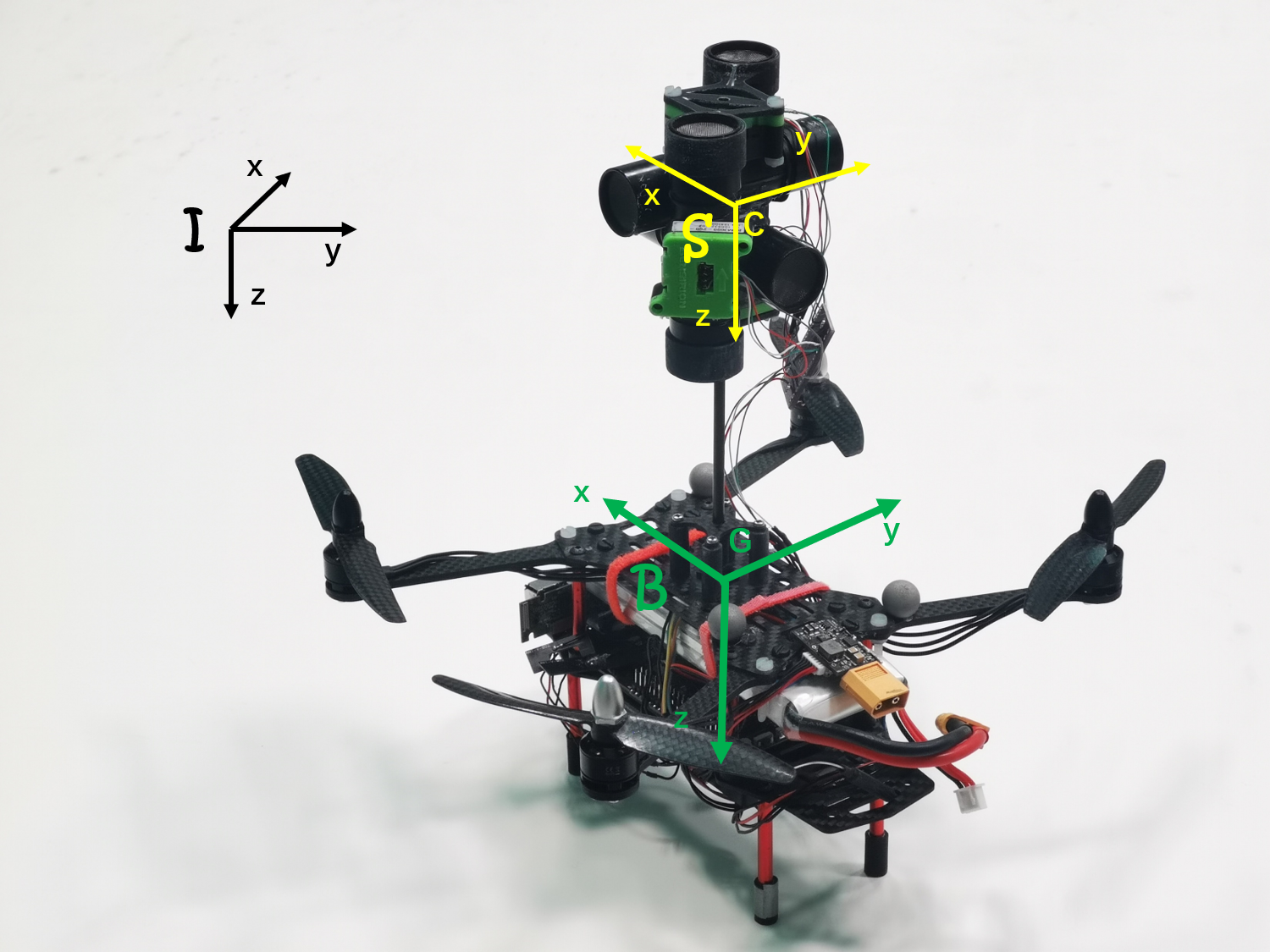}
	\caption{MAV equipped with four thermal anemometers orthogonal to each other.}
	\label{figure:frame}
\end{figure}

The thermal anemometer measures the speed of the airflow passing through the cylindrical channel \cite{Muller(2022)}. It is not straightforward that a set of three orthogonal thermal anemometers estimates the 3D velocity/air-velocity of a quad-rotor MAV. The main difficulty comes from the airflow generated by the propellers, especially in the vertical direction (in body frame) where airflow velocity is the sum of the induced velocity and MAV's velocity. In addition, measurements are noisy due to electronic or mechanical vibrations. This work builds on our previous work \cite{Wang(2022)} where we demonstrated that anemometer measurements generally exhibit homoscedasticity and introduced a neural network approach to predicting air velocity. In contrast to previous work \cite{Wang(2022)}, i) we use a symmetric anemometer configuration to improve the stability of the vertical velocity estimation, ii) we use a learning approach to estimate the flight acceleration based on aerodynamic model, iii) we employ barometer measurements to reduce the cumulative drift, and iv) we fuse the attitude estimation from onboard IMU measurements, v) in addition, we use a learning-based approach to identify whether the vehicle is in the air or on the ground to avoid the serious ground effect on measurements (especially for barometric measurements).

The parer is organized as follows. Section \ref{related work} presents related work, including data fusion strategies, different anemometers, and applications for aerodynamic system identification. Section \ref{overview} presents the system diagram of the proposed approach. Section \ref{network} describes our neural network architecture, one structure for all. Section \ref{AIO} describes the observer and the data fusion. Section \ref{result} describes the implementation details and presents the results. The conclusion and perspectives are placed in the last section.

\section{Related Work} \label{related work}

Since a single sensor cannot cope with all situations, a common odometry system employs data fusion methods \cite{Tagliabue(2021)}. The data includes multiple different sensor measurements \cite{Li(2019)}, model-based virtual sensor measurements \cite{Mur-Artal(2015)}, and constrained sensor measurements \cite{Lew(2019)}. Since low-cost acceleration suffers from severe integration drift, these methods fuse additional velocity or position measurements from other sensors. Mobile robot motion is subject to its intrinsic dynamics model \cite{Svacha(2020)}, so that the identified systems could be used in turn as virtual sensor measurements \cite{Letalenet(2020)}. In special scenarios, we get extra valuable information due to constrained sensor measurements \cite{Lew(2019)}. In this work, without exception, we made use of measurements from any on-board sensor/virtual sensor to build the odometry system, which includes 9-axis low-cost IMU, 3-axis anemometer, barometer, quad-rotor commands.

The velocity of the airflow surrounding a quad-rotor can be measured by anemometers, which could be divided into two categories: velocity anemometers (Cap/Vane anemometer, Hot-wire anemometer \cite{Simon(2022)}\cite{Simon2(2022)}, Ultrasonic anemometer \cite{Hollenbeck(2018)}), and dynamic pressure/drag anemometers (Pitot-Tube based anemometer, Hall-based anemometer \cite{Kim(2019)}\cite{Kim(2020)}). Recently, researchers have designed new anemometers based on the similar principle, which either directly convert airflow velocity into other readable physical signals \cite{Sterbing(2011)} or convert dynamic pressure into readable signals and subsequently derive airflow velocity \cite{Zahran(2018)}. In particular inspired by bionic ideas, a whisker-like sensor \cite{Tagliabue(2020)} is presented and Hall sensors are used to measure the displacement of magnetic objects caused by air drag. The characteristics of various anemometers differ greatly due to the difference in sensing principles. Micro quad-rotor UAVs are limited by payload and computing power, so we need an anemometer with features such as light weight, compactness, high sampling rate, high accuracy, no mechanical moving parts, easy to maintain and inexpensive. The SFM3000 thermal anemometer was also used , which is a 1D bidirectional anemometer and weights only 17g, with a data output rate of 2kHz, a measurement range of 200 standard liters per minute (slm) and a resolution of ±2.5\%.

Multi-rotor UAVs generate thrust and 3-axis moments through high-speed spinning propellers to change the flight attitude and flight speed. Since the propeller mass is much smaller than that of the UAV, the UAV is often considered as a mass point subjected to forces in four degrees of freedom. The propeller speed is controlled by an electronic speed controller (ESC), so the ESC commands are often used as input variables to identify the UAV dynamics system \cite{Svacha(2020)}. System identification is a fitting process, so models are often designed from experience and then approximated using filter methods to identify parameters or neural network methods \cite{Connell(2022)}. In this work, we use the same network structure for estimating acceleration as for airflow velocity estimation, since the input data are also time-series in nature.

\section{Overview} \label{overview}
\subsection{System Diagram}
\begin{figure}[t]
	\centering
	\includegraphics[scale=0.72]{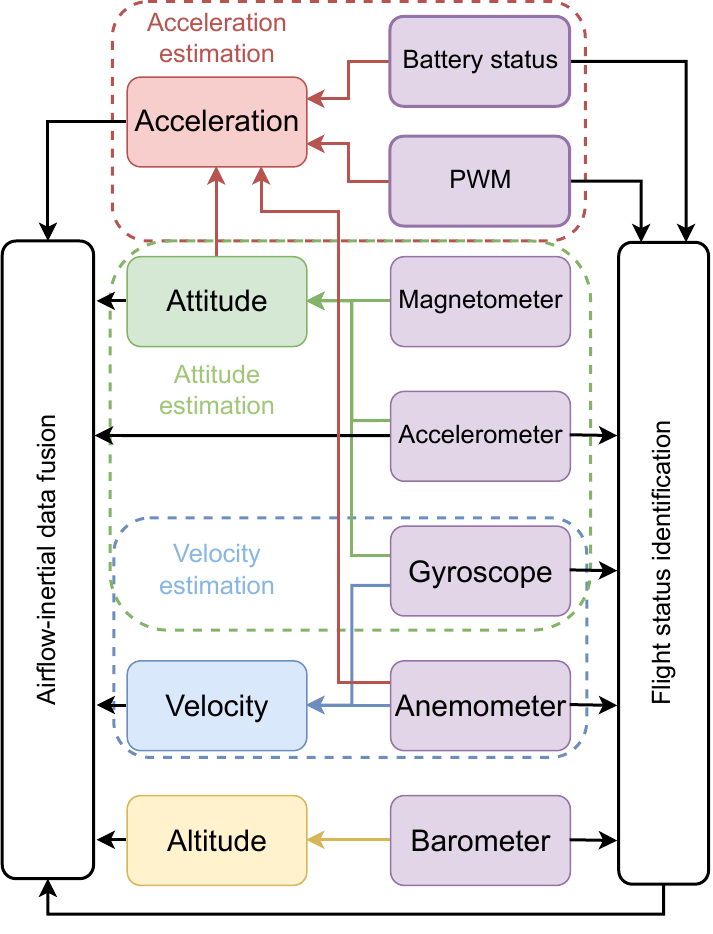}
	\caption{System diagram of the proposed approach. The purple blocks are the onboard sensors. The attitude is estimated from the 9-axis IMU. The acceleration estimator, velocity estimator and flight status identifier use similar neural networks, due to their same time-series characteristics. The airflow inertia estimator fuses all available information.}
	\label{figure:arch}
\end{figure}
The proposed approach consists of several components, shown in Fig.\ref{figure:arch}. \textbf{i) Sensor:} The battery monitor, ESC, 9-axis IMU, and barometer constitute the I/O interface of the simplest multi-rotor UAV. An additional 3-axis anemometer is added for flight speed measurement, and they together provide the input to our odometry system, shown by purple blocks. \textbf{ii) Acceleration estimator:} Leveraging machine-learning methods, an identified quad-rotor aerodynamic model is included in the acceleration estimator, which utilizes information such as: battery status (voltage, current), motor control volume, relative air velocity, and attitude. \textbf{iii) Attitude estimator:} The attitude is estimated from the 9-axis IMU measurements. \textbf{iv) Velocity estimator:} Anemometer measurements are used to estimate the vehicle's flight speed, and angular velocity measurements are used as additional inputs since the sensors are mounted at non-center-of-mass locations. \textbf{v) Altitude measurement:} Based on the pressure altitude formula, the relative altitude change during flight is measured from the barometric measurements. \textbf{vi) Flight status identifier:} Our odometry system is applicable for the full flight lifetime, so ground effects during the takeoff/landing phase are also considered. Fortunately the constraint that the flight speed is zero when the aircraft is on the ground is known, so we design a learning-based flight status identifier. \textbf{vii) AIO:} A observer based airflow inertial odometry fuses all information.

\subsection{Notation}

\begin{itemize}
	\item $\{I:{i_0},{j_0},{k_0}\}$ is a NED inertial frame.
	\item $\{B:{i},{j},{k}\}$ is a body frame (i.e., attached to the MAV), with center at the MAV's center of mass $G$. The rotation matrix from 
	$\{B\}$ to $\{I\}$ is denoted as $R$ and it satisfies the equation $\dot R = R [\omega]_{\times}$ with $\omega$
	the angular velocity vector expressed in body frame and
	$ [\cdot]_{\times}$ the skew-symmetric matrix associated with the cross product. 
	\item $\{S\}$ is a sensor frame (i.e., attached to the triaxial anemometers), with center at its center $C$. The constant rotation matrix
	from $\{S\}$ to $\{B\}$ is denoted as $R_0$ and the vector of coordinates of $C$ in the body frame $\{B\}$ is denoted
	as $\delta$.
	\item $P$, $V$, $A$ denote the coordinates, in inertial frame, of the position/velocity/acceleration vector of $G$.
	\item $V_{w}$ denotes the coordinates, in inertial frame, of the wind speed vector.
\end{itemize}

\section{Neural Network: one structure for all} \label{network}
\begin{figure}
	\centering
	\begin{subfigure}{0.16\textwidth}
		\includegraphics[width=\textwidth]{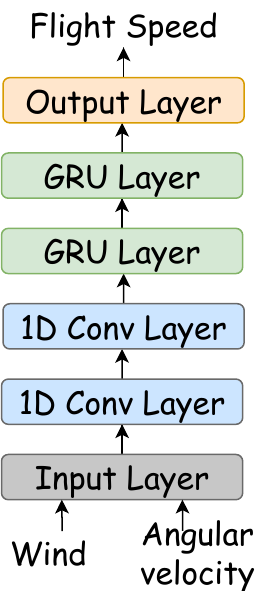}
		\caption{Velocity.}
		\label{fig:vel_arch}
	\end{subfigure}
	\hfill
	\begin{subfigure}{0.15\textwidth}
		\includegraphics[width=\textwidth]{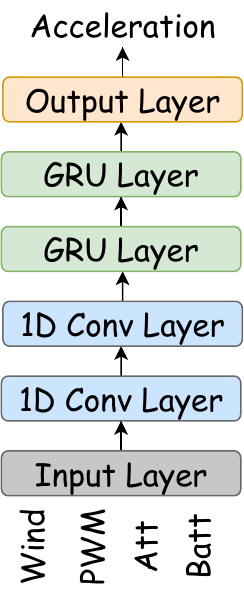}
		\caption{Acceleration.}
		\label{fig:acc_arch}
	\end{subfigure}
	\hfill
	\begin{subfigure}{0.163\textwidth}
		\includegraphics[width=\textwidth]{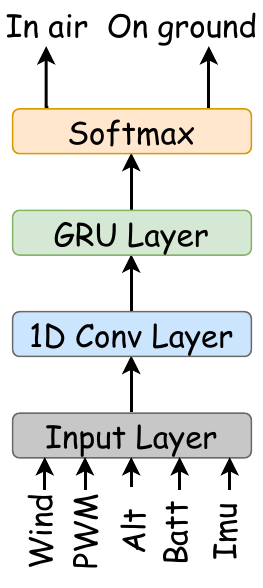}
		\caption{Flight status.}
		\label{fig:status_arch}
	\end{subfigure}
	
	\caption{Neural network architecture of the proposed approach.}
	\label{fig:arch}
\end{figure}

For state estimation, model-based or data-driven approaches are often used. A dynamical model is often designed to approximate the system state transfer process based on experience, and a filter based approach is used to estimate the dimensionless parameters in the model to estimate the state values or output based on the measurements or inputs. The data-driven approach regards the system state transfer process as a black box, and fits a set of nonlinear mapping relations so that the inputs or measurements can correspond to the states or outputs. For state estimation tasks, data-driven neural network is often used as a predictor due to its ability to approximate arbitrary nonlinear functions. The neural network architectures of the proposed approach used for all estimations in the work are shown in Fig.\ref{fig:arch}

Due to electronic noise and mechanical vibrations, the on-board sensors inevitably output measurements with significant noise. Compared to the signal, such noise is usually of high frequency and previous work \cite{Wang2(2022)} found that the variance of the noise is not always constant, but rather correlates with the flight state, so we have to look for an adaptive approach. Wavenet network \cite{Oord(2016)}, a network structure dedicated to the speech recognition domain, suggests us that convolutional neural networks (CNN) networks with a broader field of view can cope well, so in the proposed approach the CNN network is used first for extracting effective features.

CNNs are able to reduce noise based on a slice of data due to the nature of their structure, but when we want to extract information from longer slice, we need to design a larger convolution kernel thus increasing the risk of overfitting. In fact, in this work, we not only need to consider the measurement noise, but also need to extract deeper information such as the induced downwash flow rate, ground effect, dynamical model, constant parameters $R_0$ and $\delta$, etc. Sensor data have typical time-series characteristics that are very convenient for using recurrent neural networks (RNN). In order to learn the mapping relations over a long range without causing gradient explosion or gradient disappearance, gating mechanisms have been proposed. GRU networks \cite{Cho(2014)} further reduce the number of parameters based on LSTM networks \cite{Hochreiter(1997)} and thus reduce the risk of overfitting. Therefore, the GRU layer is set up following the CNN layer.

\subsection{Velocity Estimator} \label{vel}
In this section, the objective is to estimate the MAV's air-velocity $V_a=V-V_w$. As mentioned in Section \ref{introduction}, this work concerns MAV's odometry in a wind-free environment, and thus $V_w=0$. Let $V_C$ denote the coordinates, in inertial frame, of the velocity of $C$. We have $V^C=V+R(\omega\times\delta)$ and thus $$ V_a^C:=V^C-V_w=V_a+R(\omega\times\delta)$$ where $V_a^C$ denotes the air-velocity at the sensor center $C$, expressed in inertial frame. Since the anemometer measurements is in the sensor frame $\{S\}$, and the rotation matrix from sensor frame to inertial frame is $\bar{R}=RR_0$, the anemometer measurement vector is $$ \bar{V}_a^C=\bar{R}^TV_a^C=R_0^T(R^TV_a+\omega\times\delta)$$ which is equivalent to \begin{equation}
	\label{vel_model}
	V_a^{\{B\}}=R^T V_a= R_0 \bar V_a^C - \omega \times \delta
\end{equation}
The structure of the proposed learning-based solution is shown in the sub-figure Fig.\ref{fig:vel_arch}, which estimates $V_a^{\{B\}}$ from the anemometer measurements $\bar{V}_a^C$ and the gyroscope measurements $\omega$. The constant parameters $R_0$ and $\delta$ will be implicitly sensed.

The neural network is composed of two 1d-CNN layers, two GRU layers, and one fully connected layer (aka. output layer). The inputs consist of a sequence of raw anemometer measurements in sensor frame and a sequence of angular rate measurements in body frame, collected during a time window. Two layers of 1d-CNN network are used for denoising, where there is 16 filters, the kernel size is 5, the stride is 1, the activation function is ReLU \cite{Nair(2010)}, and the input/output sequences are of the same length by padding with zeros. Two layers of GRU network with 16 units are stacked with following the CNN to decouple the induced airflow from flight speed. In order to keep the network invariant w.r.t the vehicle's orientation, the output is supervised by ground truth flight velocity vector in body frame.

\subsection{Acceleration Estimation}
In this section, the objective is to predict the acceleration of the vehicle $A$ based on the ESC commands $\Omega_c$, current attitude $R$ and relative air velocity $V_a^{\{B\}}$. The MAV's aerodynamic model is approximated by a neural network. According to the previous work \cite{Letalenet(2020)} on position dynamics, Newton equation yields:
\begin{equation}
	\label{acc_model}
	mA = mg - R(T(V_a,\eta,\Omega) + F_a(V_a))
\end{equation}
where the aerodynamic drag is defined as $F_a\varpropto|V_a|V^{\{B\}}_a$. $T$ denotes the combined force generated by the spinning propeller, including the trust perpendicular to the rotor plane and the horizontal forces parallel to the rotor plane, according to the Blade Element Momentum Theory \cite{Bramwell(2001)}. 

The propeller spinning rate is controlled by the ESC with a rotor's dynamic equation of $ \dot{\Omega}=\frac{1}{\tau}(\Omega_c-\Omega)$. The desired speed of $\Omega_c$ is proportional to ESC command. Therefore, according to the theories there exists a mapping between accelerations and inputs. Although the rotor dynamics is a time-delayed system, the MAV motor response time is small. Since the neural network system observes data over a period of time at a time, which is much larger than the motor response time, it is not affected by system delays. As mentioned in Section \ref{vel}, the relative air velocity $V^{\{B\}}_a$ is estimated by the velocity estimator, by decoupling the induced velocity $\eta$ from the measurements. The acceleration estimator has the same network structure as the velocity estimator, as shown in the sub-figure Fig.\ref{fig:acc_arch}.

The inputs consist of a sequence of raw anemometer measurements, ESC commands, MAV's attitude estimated by an observer, and battery statuses (voltage and current), collected during a time window. Two layers of 1d-CNN are configured with 5 filters, the kernel size is 12, the activation function is ReLU, and the input/output are of the same length. Two layers of GRU network have 12 units, and the second layer return only the last output. The output is supervised by ground truth flight acceleration in inertial frame.

\subsection{Flight Status Identification}
The odometry system usually works over the whole flight life and therefore the pre-takeoff phase and the post-landing phase should be considered. However, the thrust of the UAV is generated from the downwash flow. When the vehicle is on the ground, the constraint of ground support force keeps the vehicle stationary. During the vehicle's departure from the ground, the constraint of the downwash flow by the ground in close proximity can cause local static pressure to rise and local airflow to be disturbed. Both of the cases are significantly different from that when the vehicle is in the air. Therefore, in this section, the objective is to classify the flight status by building a learning-based identifier.

The inputs consist of the mean and difference of two vertical anemometer measurements (there are two sensor mounted along vertical direction as shown in Fig.\ref{figure:frame}), the mean of ESC commands (normalized), battery status (voltage and current, normalized), the altitude derived from barometric data (shifted to ensure that the first item is zero), and mean of acceleration and angular rate (normalized). The first three sets of data have similarity in that the mean value is zero when on the ground and the value is around a non-zero constant when in the air. During the takeoff and landing phases, the altitude values show significant abnormal data, and the IMU data vibration characteristics can change rapidly. All the information can be used to recognize flight status, and they all have time-series characteristics, so the same network architecture is used, shown in sub-figure Fig.\ref{fig:status_arch}. One layer of 1d-CNN is configured with 4 filters, the kernel is 5, the activation function is ReLU, and the "same" padding is used. One GRU layer has 6 units, and return only the last output. One output layer with softmax activation function is used at the end for classification.

\section{Airflow-Inertial Odometry} \label{AIO}
Since the sensor sampling frequencies are usually different, the measurements may be of different order, and the results predicted by the neural network may have small bias, this section presents an observer-based approach by fusing the various outputs and the predicted results of the neural network are treated as sensor measurements.
\subsection{Attitude Estimation}
A typical IMU includes a 3-axis accelerometer, 3-axis gyroscope, 3-axis magnetometer, which have the measurement model with biases $b_*$:
\begin{equation}
	\label{imu_model}
	\left \{
	\begin{aligned}
		a_m &= R^T(\dot{v}-gk_0)+b_a \in \{B\} \\
		g_m &= \omega+b_g \in \{B\} \\
		m_m &= R^Tm+b_m \in \{B\}
	\end{aligned}
	\right.
\end{equation}
where $\dot{v}$, $\omega$ are the real value, and $m$ is the Earth's magnetic field vector.
According to the observation from gravity and Earth's magnetic field vector, a simple attitude estimator \cite{Mahony(2012)} can be:
\begin{equation}
	\label{att_model}
	\left \{
	\begin{aligned}
		&\dot{\hat{R}} := \hat{R}[g_m-\hat{b}_g]_\times - \gamma \\
		&\dot{\hat{b}}_g := \alpha_8a \\
		&\gamma := [\frac{\alpha_9}{g}((\hat{R}^Tk_0)\times (\bar{a}_m-\hat{\dot{v}}))+\frac{\alpha_{10}}{|m|^2}((\hat{R}^Tm)\times \bar{m}_m)]_\times
	\end{aligned}
	\right.
\end{equation}
where $\bar{a}_m$, $\bar{m}_m$ are the mean of several sequential sampling, and $\hat{\dot{v}}$ is estimated acceleration.

\subsection{Altitude Measurement}
Since the air temperature is linear with altitude in the troposphere at the altitude below 10 km (eq: $T = T_1 + L(z-h_1)$), according to the ideal gas law (eq: $pVM = mRT$) and the hydrostatic equation (eq: $dp = -\rho gdz$), we have the pressure-altitude equation:
\begin{equation}
	\label{alt_model}
	h = h_1 + \frac{T_1}{L}\left[(\frac{P}{P_1})^{-\frac{RL}{gM}}-1\right]
\end{equation}
where $h_1, T_1, P_1$ are altitude/temperature/pressure at the reference level (e.g sea level), and $R= 8.31432\left[\frac{Nm}{molK}\right]$, $L= -0.0065\left[\frac{K}{m}\right]$, $M= 0.0289644\left[\frac{kg}{mol}\right]$, $g= 9.80665\left[\frac{m}{s^2}\right]$ are universal gas constant, standard temperature lapse rate, molar mass of Earth’s air, gravity constant, respectively.

\subsection{Data Fusion}
Anemometer, IMU, ESC, and barometer are the main sources of information acquisition in this work. Prior to this section, the velocity estimator estimates the flight speed in the body frame using anemometer measurements, the acceleration estimator estimates the acceleration in the inertial frame using ESC commands, the IMU is used to estimate the attitude as well as provide acceleration measurements, and the relative altitude is calculated based on barometer measurements. 
When the vehicle is on the ground, the vehicle speed is close to zero due to the ground constraint. We have then an observer in Eq.\ref{oberser_on_ground}.
Since errors in measurements and neural network prediction always exist, we modeling them with time-invariant bias in the observer. According to the experimental data in the air, the author found that the mean of barometer measurement bias varies slowly, and thus the barometer bias is set as a first-order equation. We have then an observer in Eq.\ref{oberser_in_air}. $h, \hat{b}_b$ denote the altitude measurements and estimated bias, $v_w, \hat{b}_w$ denote the output of velocity estimator and estimated bias, $a_m, \hat{b}_a$ denote the accelerometer measurements and estimated bias, and $a_p$ denotes the output of acceleration estimator. $k_*, \alpha, \beta$ are non-negative gains. $\hat{\cdot}$ denotes the estimation of a state in the observer.

When the vehicle is on the ground, the observer is
\begin{equation}
	\label{oberser_on_ground}
	\left \{
	\begin{aligned}
		&\dot{\hat{P}}=\hat{V}\\
		&\dot{\hat{b}}_a=k_0\hat{V}\\
		&\dot{\hat{V}}=\hat{A}-k_1\hat{V}\\
		&\hat{A}=g+Ra_m-\hat{b}_a\\
	\end{aligned}
	\right.
\end{equation}
when the vehicle is in the air, the observer is
\begin{equation}
	\label{oberser_in_air}
	\left \{
	\begin{aligned}
		&\dot{\hat{P}}=\hat{V} -k_2(\hat{P}-h+\hat{b}_b)\\
		&\dot{\hat{V}}=\hat{A}-k_3(\hat{V}-Rv_w+\hat{b}_w)\\
		&\hat{A}=\alpha(g+Ra_m-\hat{b}_a)+(1-\alpha)a_p\\
		&\dot{\hat{b}}_a=k_4(\hat{V}-Rv_w+\hat{b}_w)+k_5(\hat{P}-h+\hat{b}_b)\\
		&\dot{\hat{b}}_w=-k_6(\hat{V}-Rv_w+\hat{b}_w)\\
		&\ddot{\hat{b}}_b=-\beta k_5(\hat{P}-h+\hat{b}_b)\\
	\end{aligned}
	\right.
\end{equation}
%

\section{Implementation \& Results} \label{result}
\subsection{Experimental Platform}
We collect data on the custom MAV flight platform of Figure \ref{figure:frame} with a frame of 15x20x20cm and weight about 500g. The MAV is equipped with Pixhawk 4 autopilot (STM32f7), and Khadas Vim 3 SBC (6-core ARM 2.4GHz). The open source flight control system PX4 runs on Pixhawk 4 and contains a 4-loops cascade PID controller, an Extended Kalman Filter (EKF) based state estimator, and uses the mavlink protocol to communicate with Vim 3 via the serial port. All components run in a ROS system on Vim 3 and communicate with an ground station laptop via WiFi5. 

\subsection{Data Collection}
The anemometers, thermal mass flow sensor SFM3000 from Sensirion AG, run at 500Hz, and communicate with Pixhawk 4 via high speed I2C interface. The IMU data and the barometric data are collected from Pixhawk 4 which is equipped with two low-cost high sampling rate accelerometer and gyroscope, one magnetometer, and one barometer. We collect IMU data through low-pass filter, anemometer data, attitude at 400 Hz, barometer data at 200 Hz, battery status data at 100 Hz, and ESC commands at 800 Hz. The ground truth position is provided by the MoCap system expressed in MoCap frame at 120 Hz. To produce supervised values for machine learning, we use a sliding average polynomial interpolation fitting method to obtain values with high frequencies. The polynomial fitting method is also used to smooth the ground truth position curve and acquire the ground truth velocity and acceleration at the same time. The EKF inside PX4 system utilize only the IMU, barometer data (without MoCap data), and estimate the vehicle's attitude. In order to obtain the ground truth, finally used, in the inertial frame, we transformed the orientation of the MoCap frame according to the attitude provided by the EKF. The horizontal plane of the MoCap frame is essentially the same as the inertial frame and is therefore not transformed.

\subsection{Neural Network Training}
We collect four sets of manual random flight data with each flight lasting about three minutes. Two sets of data are used as the training set, one as the validation set, and one as the test set. As described in the Section \ref{network}, three networks are built, a velocity estimator, an acceleration estimator, and a flight status identifier. The batches of sequences are created by sliding a time window of 1s, a time window of 0.5s, and a time window of 0.5s, respectively. The all batch sizes have been set to 512, and we use the Adam with Triangular Cyclic Learning Rate (CLR) as the optimizer, which periodically increases and decreases the learning rate during the training. Early stopping is also used to avoid over-fitting.

\subsection{Learning-based Estimator Results}
\begin{figure}
	\centering
	\begin{subfigure}{0.5\textwidth}
		\includegraphics[width=\textwidth]{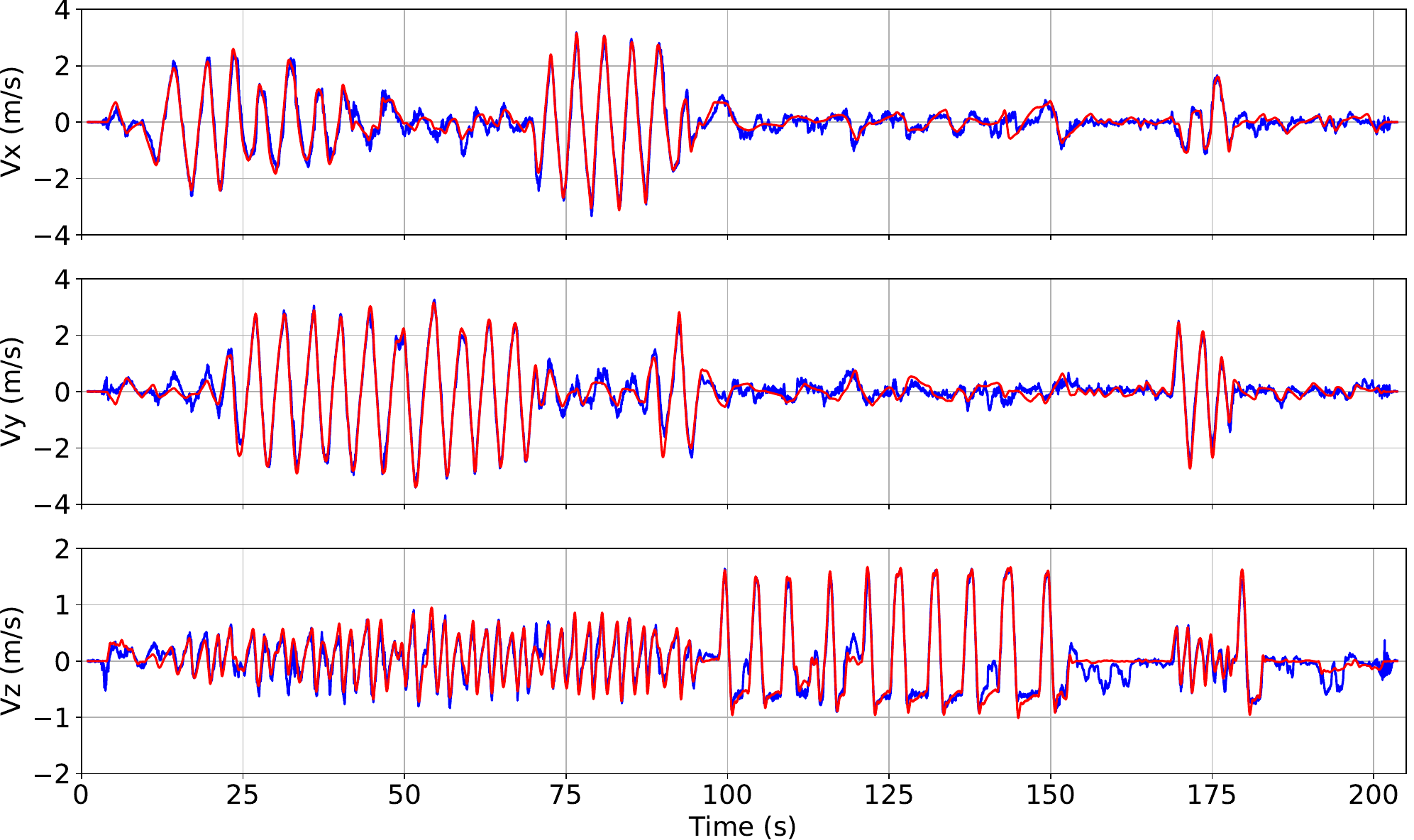}  
		\caption{Velocity prediction (blue) verse Ground truth (red) in body frame.}
		\label{fig:vel_comparison}
	\end{subfigure}
	\hfill
	\begin{subfigure}{0.5\textwidth}
		\includegraphics[width=\textwidth]{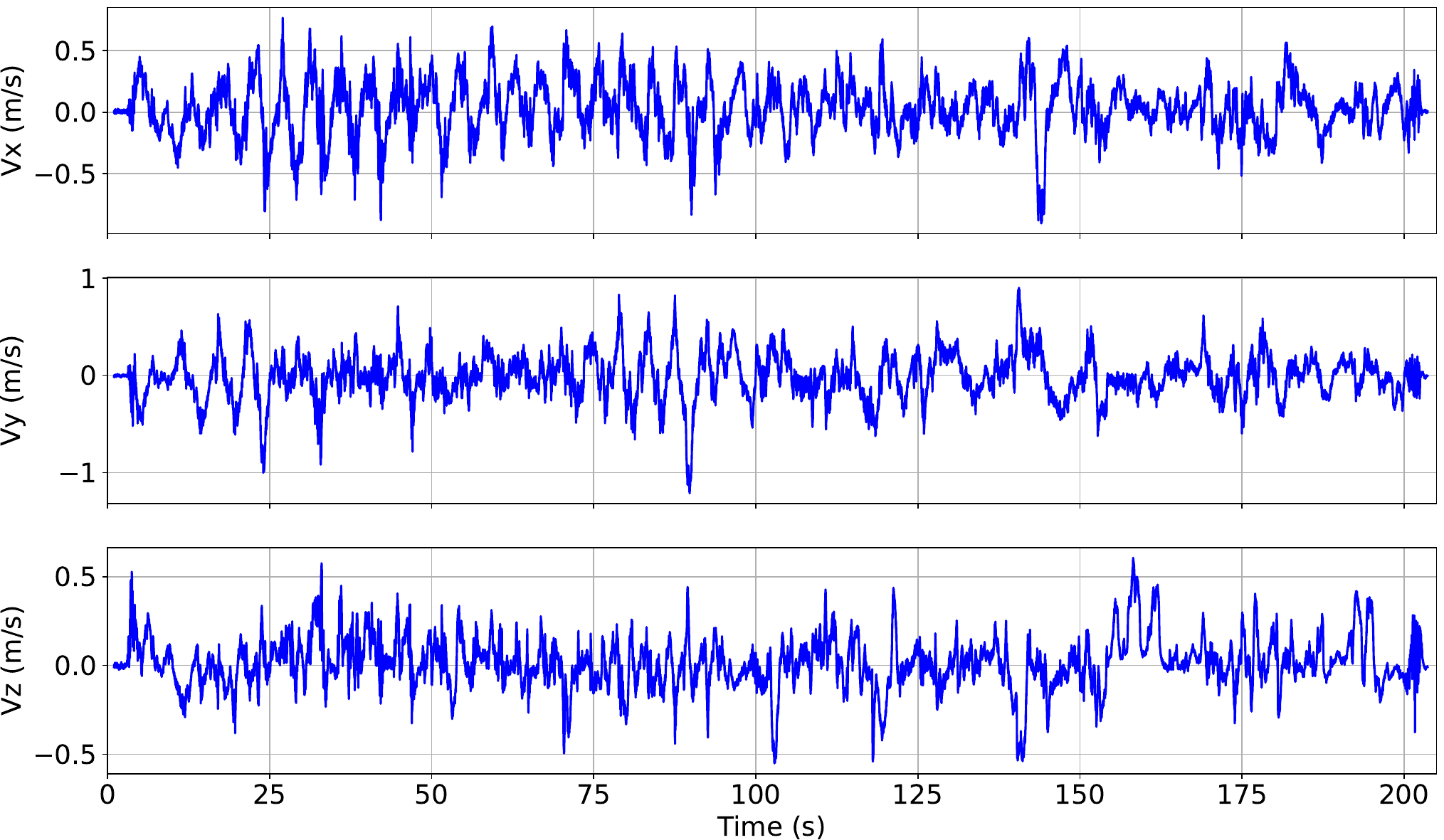}  
		\caption{Velocity prediction error in body frame.}
		\label{fig:vel_error}
	\end{subfigure}
	\hfill
	\begin{subfigure}{0.5\textwidth}
		\includegraphics[width=\textwidth]{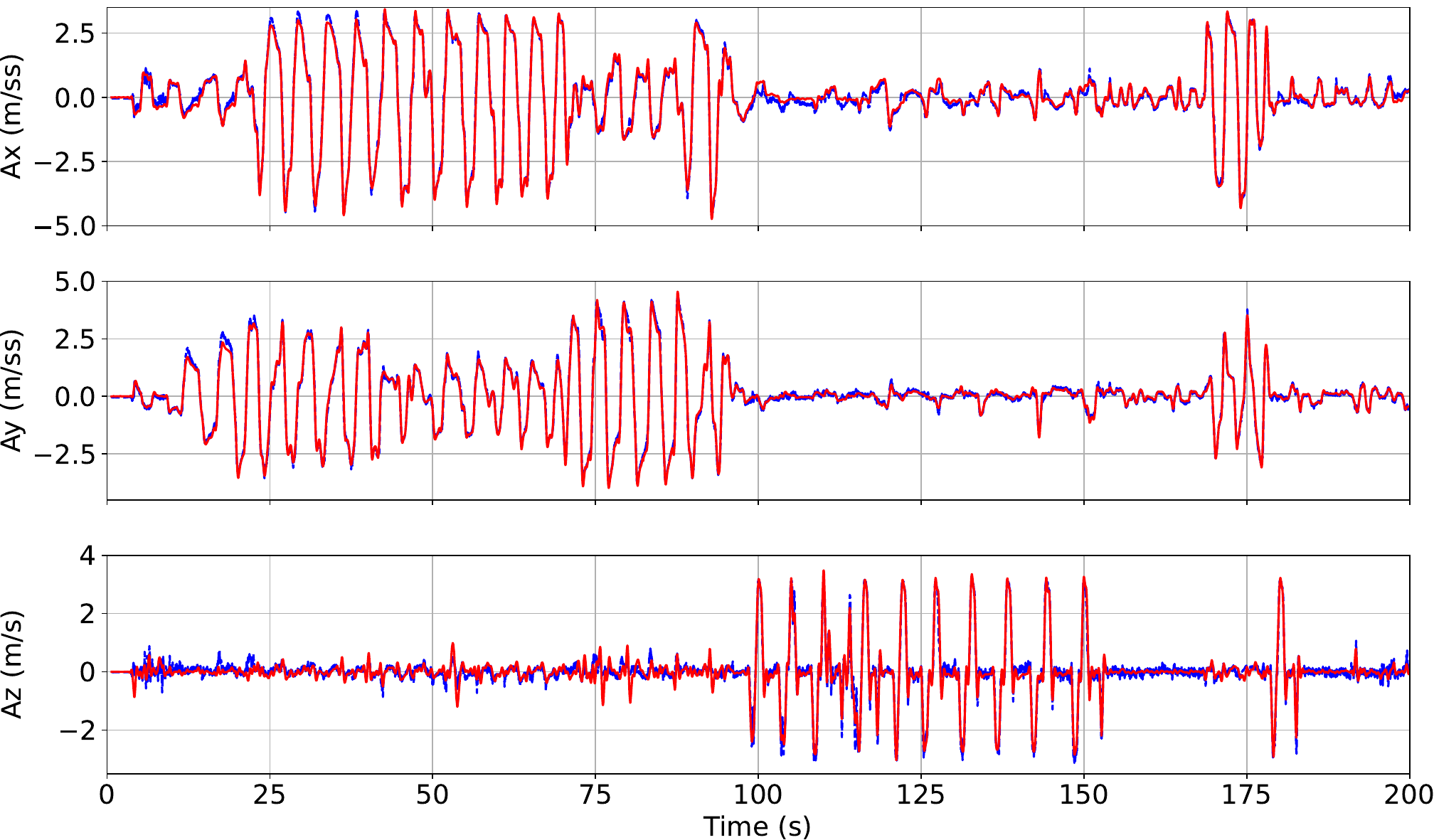}  
		\caption{Acceleration prediction (blue) verse Ground truth (red) in inertial frame.}
		\label{fig:acc_comparison3}
	\end{subfigure}
	\hfill
	\begin{subfigure}{0.5\textwidth}
		\includegraphics[width=\textwidth]{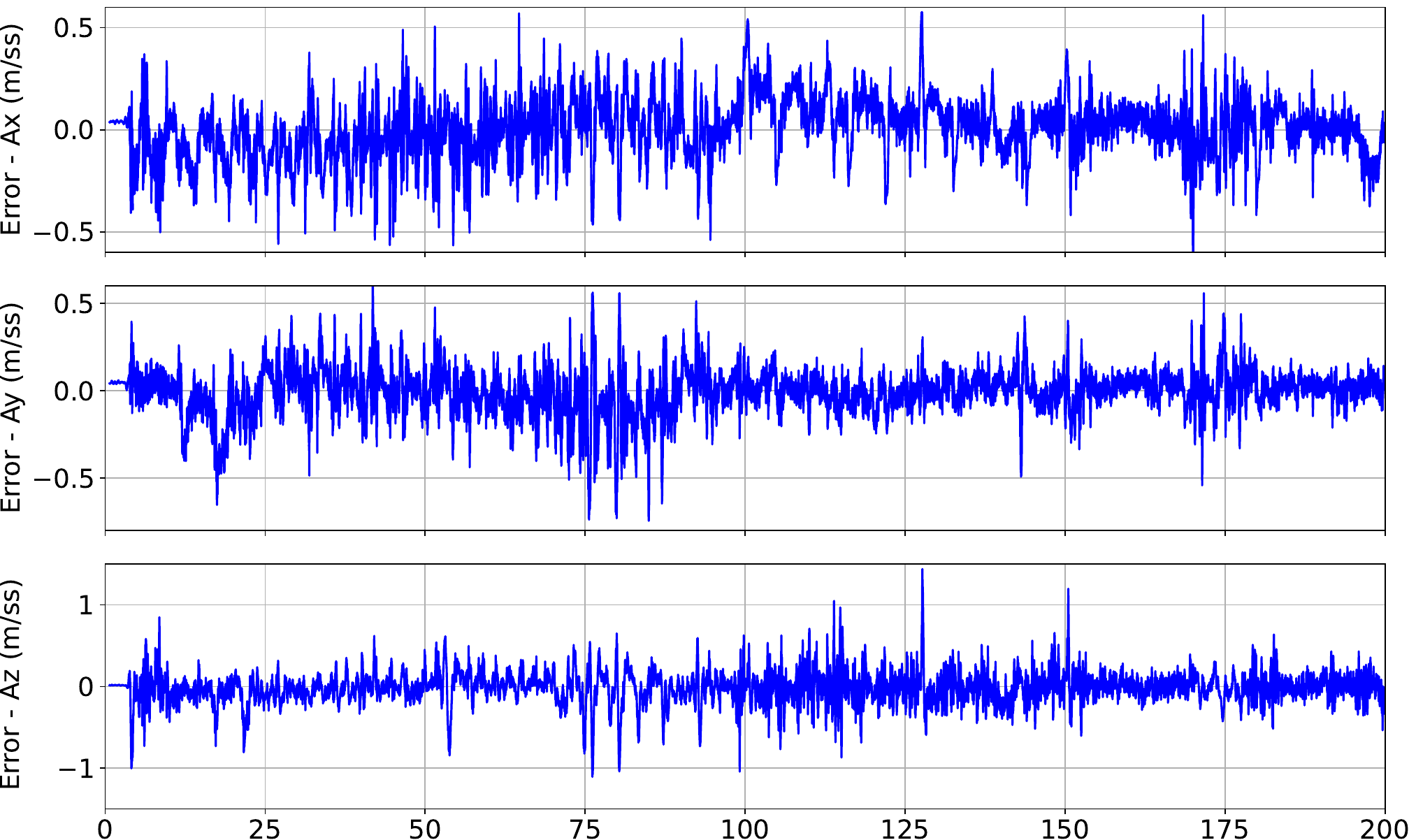} 
		\caption{Acceleration prediction error in inertial frame.}
		\label{fig:acc_error3}
	\end{subfigure}
\end{figure}
\begin{figure}
	\ContinuedFloat
	\centering
	\begin{subfigure}{0.5\textwidth}
		\includegraphics[width=\textwidth]{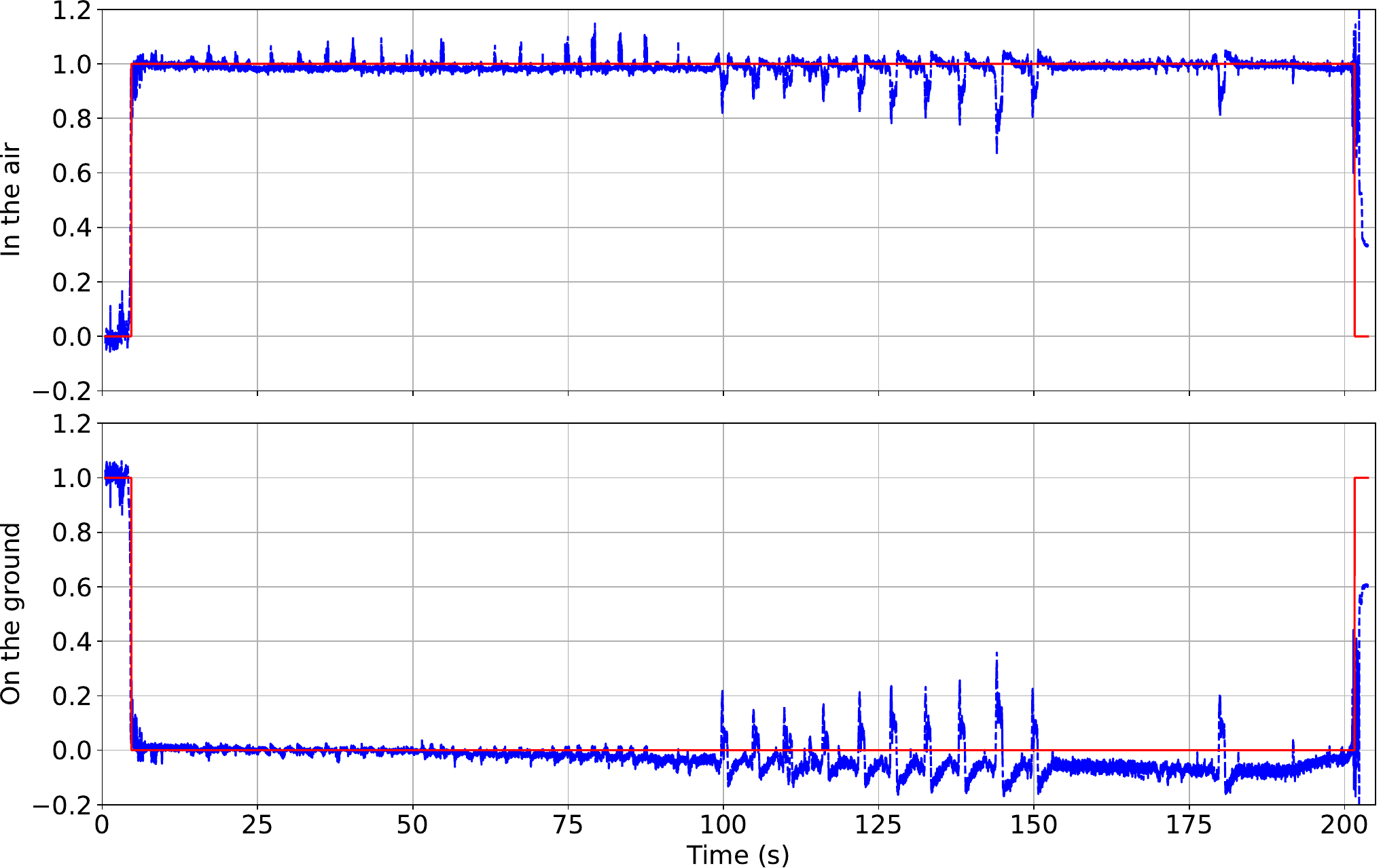} 
		\caption{Flight status identification before calculating the cross-entropy.}
		\label{fig:flight_status}
	\end{subfigure}
	
	\caption{Neural network predictions.}
	\label{fig:prediction}
\end{figure}
The Fig.\ref{fig:prediction} shows our GRU-based neural network prediction results on the validation set, including the estimation of 3D velocity in the body frame by the velocity estimator versus the ground truth and the estimation error, the estimation of 3D acceleration in the inertial frame by the acceleration estimator versus the ground truth and the estimation error, and the identification results of the flight status, respectively. Note that the horizontal time axis of the plots cover the full flight life including takeoff and landing from the ground.

To overcome the effects of gravity, the aerial vehicle requires a downwash-induced velocity, which causes the measured airflow velocity to differ from the flight velocity. The blue curve in the sub-figure Fig.\ref{fig:vel_comparison} is the neural network prediction while the red curve is the ground truth. The blue curve in the three subplots always overlaps the red curve with an error of essentially no more than $0.5$ m/s, which means that the neural network is able to implicitly estimate and decouple the current induced wind speed based on a time-slice of data.

When on the ground, the vehicle is constrained by ground support force, while when in the air, the thrust is entirely generated by the propeller, so that the ESC commands change significantly during the takeoff and landing phases. As with the velocity prediction, the acceleration prediction shown in the sub-figure Fig.\ref{fig:acc_comparison3} is able to overlap with the ground truth still during the takeoff and landing, which means that the neural network is able to recognize the flight status implicitly during the prediction and it is evidenced by the flight status identification shown in the sub-figure Fig.\ref{fig:flight_status}.

\subsection{Data Fusion Results}
\begin{figure}
	\centering
	\begin{subfigure}{0.5\textwidth}
		\includegraphics[width=\textwidth]{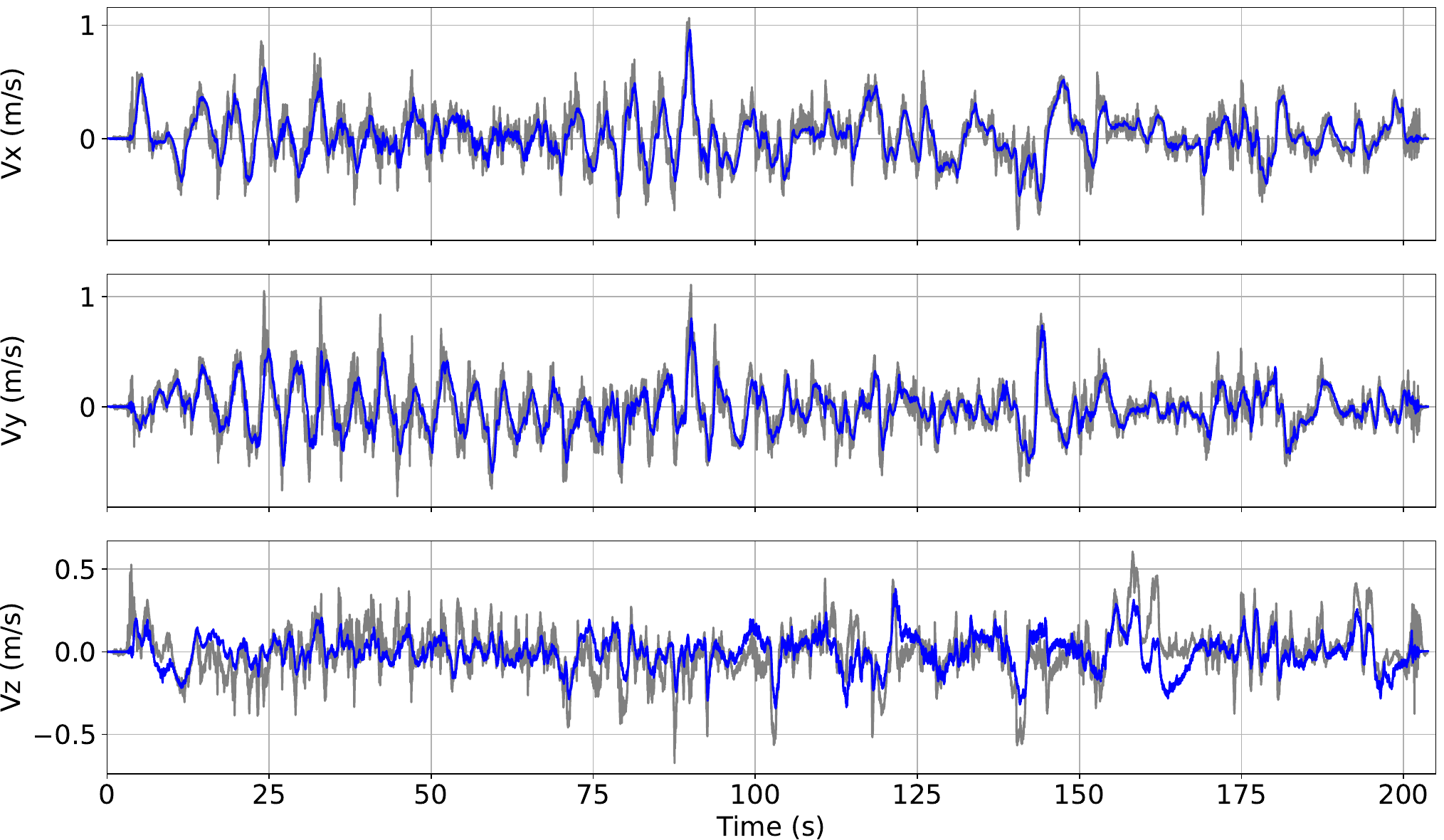}  
		\caption{Velocity estimation error in inertial frame, neural network only (gray) verse data fusion (blue).}
		\label{fig:velocity}
	\end{subfigure}
	\hfill
	\begin{subfigure}{0.5\textwidth}
		\includegraphics[width=\textwidth]{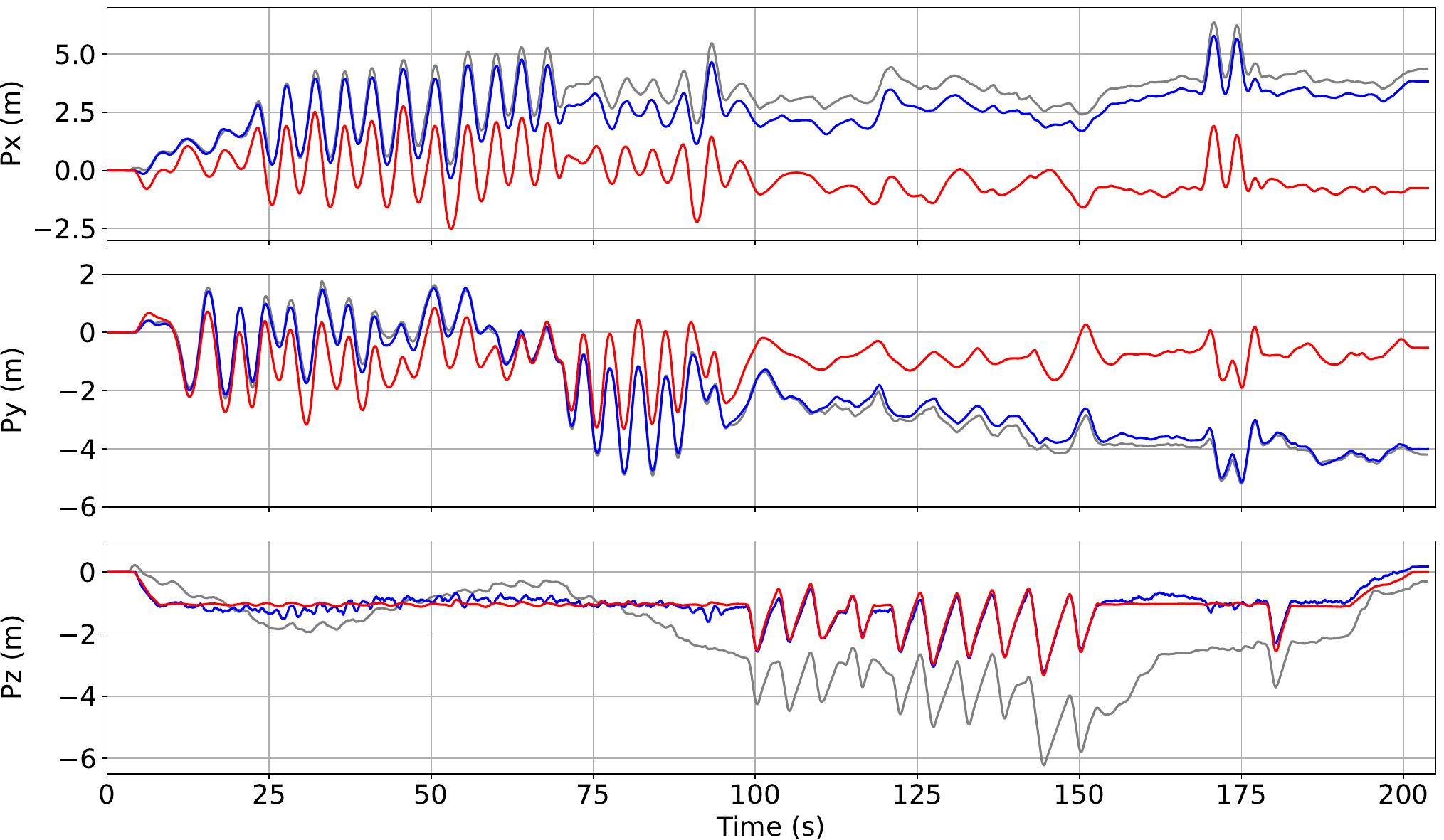}  
		\caption{Position estimation in inertial frame, neural network only (gray) \& data fusion (blue) verse Ground truth (red).}
		\label{fig:position}
	\end{subfigure}
	\hfill
	\begin{subfigure}{0.5\textwidth}
		\includegraphics[width=\textwidth]{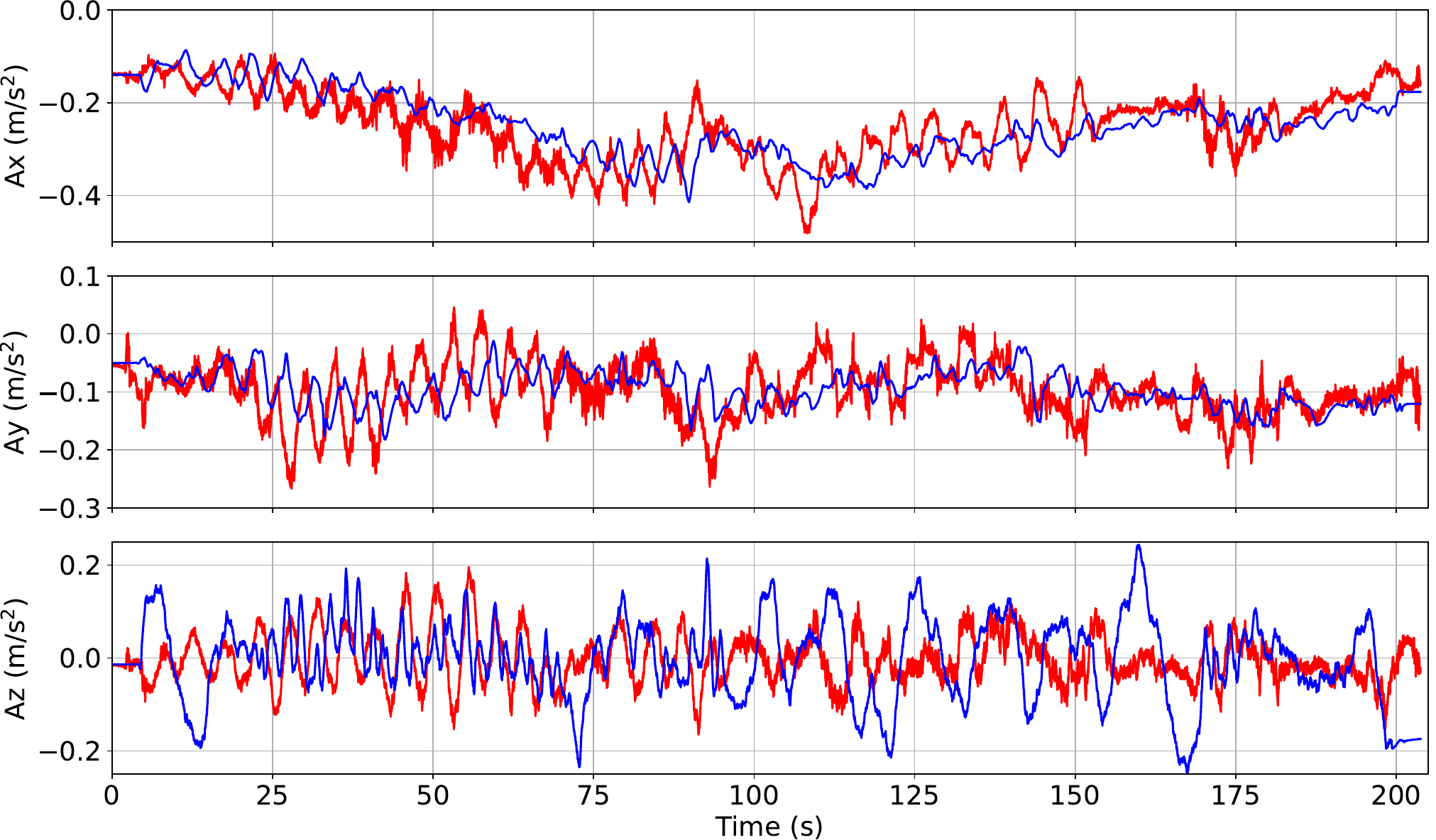}  
		\caption{Acceleration bias estimation (blue) verse Ground truth (red).}
		\label{fig:accBias}
	\end{subfigure}
	\hfill
	\begin{subfigure}{0.5\textwidth}
		\includegraphics[width=\textwidth]{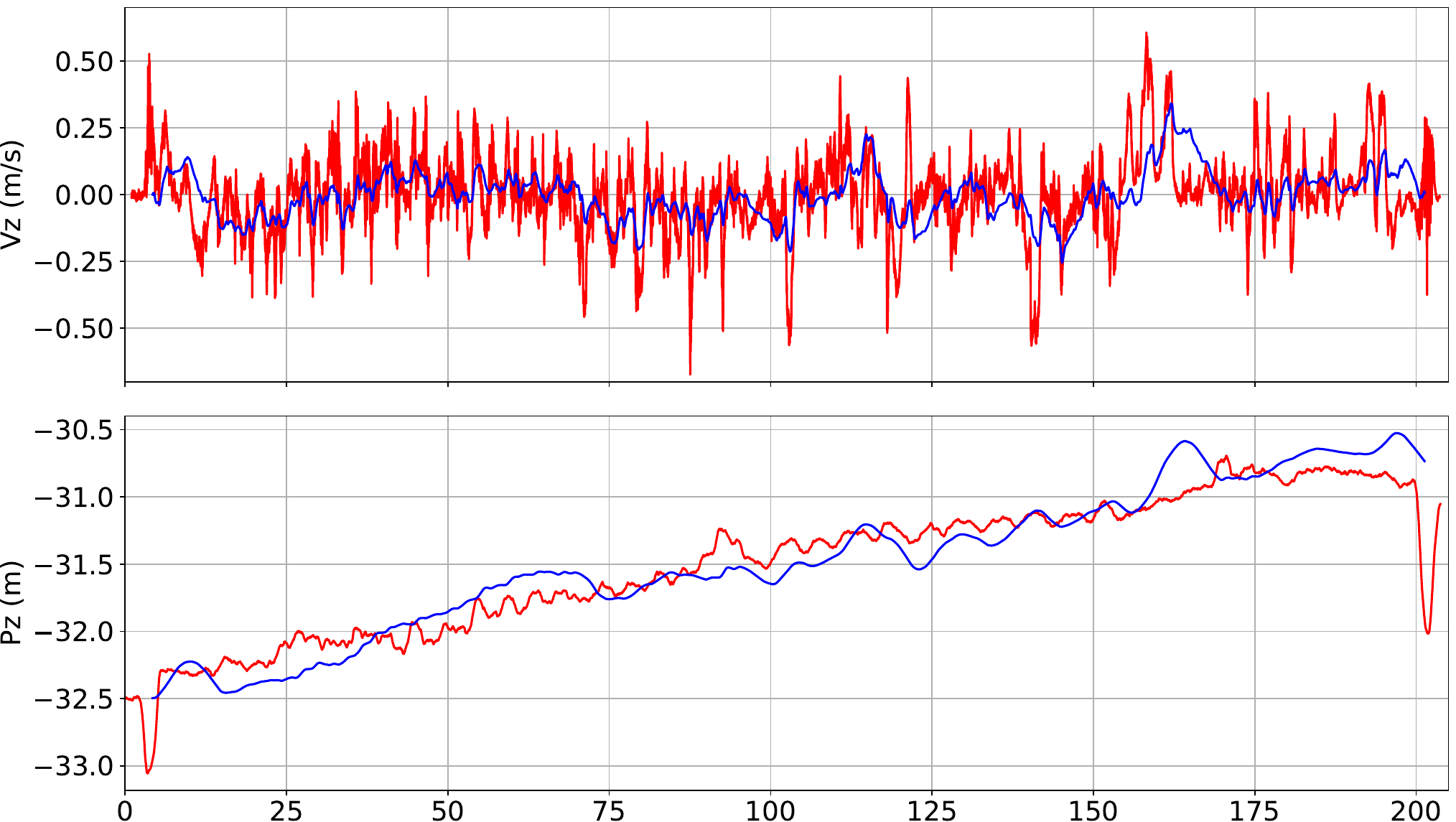} 
		\caption{Velocity estimator prediction bias in z-axis or Altitude measurement bias (blue) verse Ground truth (red).}
		\label{fig:rest}
	\end{subfigure}
	
	\caption{Data fusion results.}
	\label{fig:fusion}
\end{figure}
\begin{figure}
	\centering
	\begin{subfigure}{0.37\textwidth}
		\includegraphics[width=\textwidth]{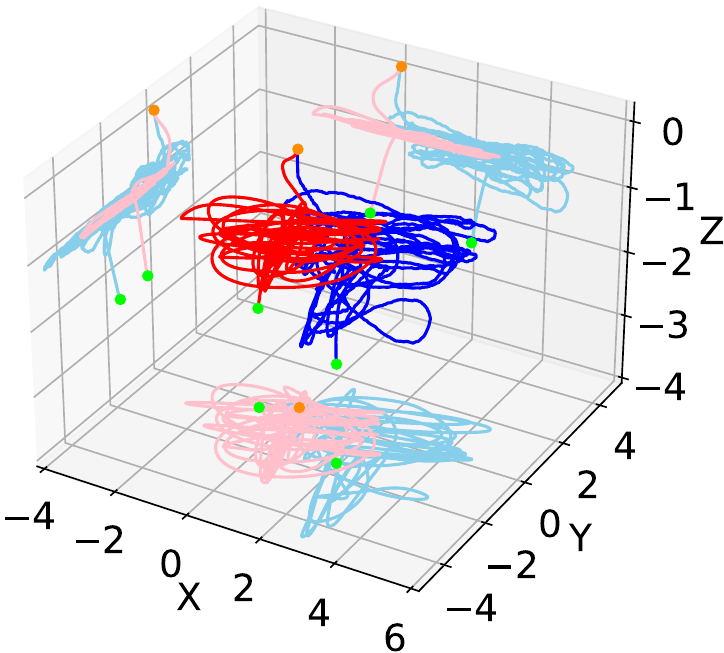}
		\caption{Flight: 0-100s.}
		\label{fig:traj_f}
	\end{subfigure}
	\hfill
	\begin{subfigure}{0.37\textwidth}
		\includegraphics[width=\textwidth]{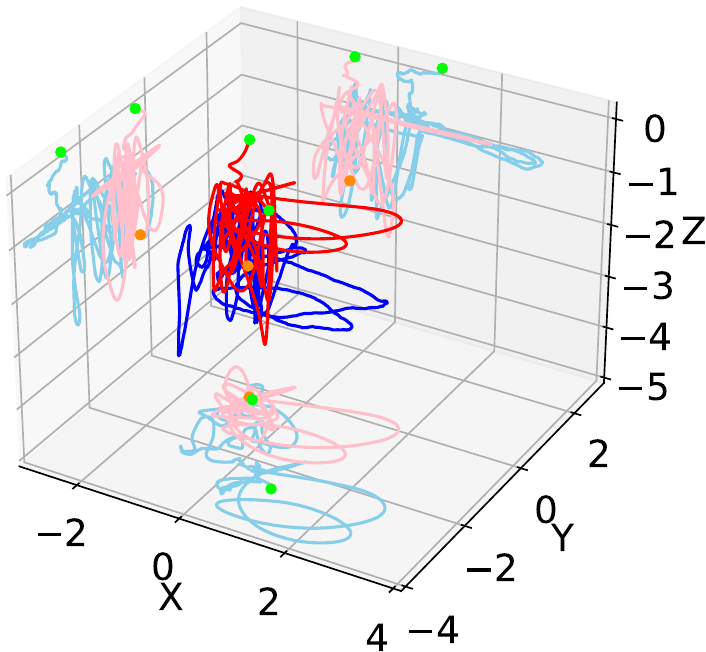}
		\caption{Flight: 100s-203s.}
		\label{fig:traj_b}
	\end{subfigure}
	
	\caption{Full flight odometry trajectory (blue) vs. the ground truth (red), from start point (orange) to the end point (green).}
	\label{fig:traj}
\end{figure}

 
\begin{table}[]
	\centering
	\begin{tabular}{||p{0.55cm} | p{0.4cm} | p{0.55cm} | p{0.5cm} | p{0.7cm} | p{0.7cm} | p{0.7cm} | p{0.25cm} | p{0.25cm}||} 
		\hline
		$k_0$ & $k_1$ & $k_2$ & $k_3$ & $k_4$ & $k_5$ & $k_6$ & $\alpha$ & $\beta$\\
		\hline
		$0.001E$ & $0.9E$ & $0.01F$ & $0.01E$ & $0.001E$ & $0.005F$ & $0.005F$ & $0$ & $0.4$ \\
		\hline
	\end{tabular}
	\caption{Non-negative factors, with $E=(\begin{smallmatrix}
			1 & & \\
			& 1 & \\
			& & 1
		\end{smallmatrix})$, $F=(\begin{smallmatrix}
			0 & & \\
			& 0 & \\
			& & 1
		\end{smallmatrix})$ }
	\label{table}
\end{table}
The non-negative gains $k_*$ and parameters $\alpha$, $\beta$ employed in Eq.\ref{oberser_in_air} and Eq.\ref{oberser_on_ground} are recorded in the Table \ref{table}. The sub-figure Fig.\ref{fig:velocity} illustrates the velocity estimation error by neural network only or by data fusion observer in inertial frame. The blue curve with the data fusion is significantly smoother and has less noise, but the velocity bias cannot be eliminated. The sub-figure Fig.\ref{fig:position} illustrates the position from integration of neural network only velocity prediction and the position estimated by the observe verse the ground truth in inertial frame. The integration drift of the vertical velocity from the velocity estimator is significantly improved since the barometer is able to measure the relative altitude, while the horizontal position cannot be corrected since either the accelerometer measurement, the acceleration estimator, or the velocity estimator has bias in the horizontal measurement. We do not then introduce bias for the acceleration estimator output and the velocity estimator horizontal velocity prediction in the observer. The sub-figures Fig.\ref{fig:accBias} and Fig.\ref{fig:rest} illustrates the estimated accelerometer measurement bias, the estimated velocity estimator output bias in vertical, and the estimated altitude measurement bias, respectively. Fig.\ref{fig:traj} illustrates the estimated full flight odometry trajectory verse the ground truth trajectory from MoCap. The total position drift is only 5.7m for 203s manual random flight.

\section{CONCLUSIONS}
In this work, we created a neural network structure specifically for sensor data processing with time-series characteristics to estimate the flight speed, acceleration, status of the vehicle based on sensor measurements. Combining 9-axis IMU and barometer, an observer performs data fusion. The observer effectively estimates the measurement bias and compensates for the integration drift of the odometry in the vertical direction. Although the odometry does not improve the integration drift in the horizontal plane due to the lack of horizontal landmark observation, this performance is sufficient to effectively address environments where GPS and vision are not available. Since authors recorded the attitude rather than magnetometer measurements when collecting the data, no learning-based approach was attempted to estimate the attitude, which will be performed in the future. MEMS laser can be used for reducing horizontal integration drift.

\addtolength{\textheight}{-0cm}   




\section*{ACKNOWLEDGMENT}
The first author is funded by the China Scholarship Council (CSC) from the Ministry of Education of P.R.C.\\
The authors want to thank Dr. Jorge ARIZAGA and Dr. Pedro CASTILLO from Heudiasyc at UTC, Compiegne, France, for giving them access to the Heudiasyc flying arena and providing support for the manual flight tests exploited in this paper.\\
The authors want to thank Prometheus\footnote{\url{https://github.com/amov-lab/Prometheus}} platform support.


\end{document}